\documentclass[journal]{IEEEtran}
\usepackage{graphicx}
\usepackage{authblk}

\ifCLASSINFOpdf
\else
\fi
\begin{document}
%
\title{WaveTouch: Active Tactile Sensing Using Vibro-Feedback for Classification of Variable Stiffness and Infill Density Objects}
%
%
%

\author[1]{Danissa Sandykbayeva}
\author[2]{Valeriya Kostyukova}
\author[3]{Aditya Shekhar Nittala}
\author[4]{Zhanat Kappassov}
\author[5]{Bakhtiyar Orazbayev}

\affil[1,3]{Department of Computer Science, University of Calgary}
\affil[2,4]{Department of Robotics, Nazarbayev University}
\affil[5]{Department of Physics, Nazarbayev University}

\maketitle

\begin{abstract}
The perception and recognition of the surroundings is one of the essential tasks for a robot. With preliminary knowledge about a target object, it can perform various manipulation tasks such as rolling motion, palpation, and force control. Minimizing possible damage to the sensing system and testing objects during manipulation are significant concerns that persist in existing research solutions. To address this need, we designed a new type of tactile sensor based on the active vibro-feedback for object stiffness classification. With this approach, the classification can be performed during the gripping process, enabling the robot to quickly estimate the appropriate level of gripping force required to avoid damaging or dropping the object. This contrasts with passive vibration sensing, which requires to be triggered by object movement and is often inefficient for establishing a secure grip. The main idea is to observe the received changes in artificially injected vibrations that propagate through objects with different physical properties and molecular structures. The experiments with soft subjects demonstrated higher absorption of the received vibrations, while the opposite is true for the rigid subjects that not only demonstrated low absorption but also enhancement of the vibration signal. 

\end{abstract}


%
\IEEEpeerreviewmaketitle

\section{Introduction}
\IEEEPARstart{V}{ibrations} are used in the majority of tactile sensing research projects due to their natural presence in real-life phenomena. For example, humans can detect slippage of gripped objects through significant changes in vibration patterns resulting from frictional forces. Inspired by such situations, many studies employed vibration response to gather knowledge about the physical properties of an object \cite{gathmann2020wearable}, \cite{muramatsu2012perception}. 

Our work suggest active vibration sensors that can classify objects according to their stiffness. Incorporating these sensors to a robotic system might facilitate various manipulation tasks such as rolling motion and force control. With active tactile feedback, we can minimize interaction between the testing object and the grasping system, thus, increasing the lifetime of the sensors and decreasing the possibility of the object's damage or slippage. This is because all the necessary data can be collected while grip with minimal exerted force while the subject stays at rest, unlike other methods involving palpation or rolling.

\begin{figure}[ht]
    \centering
    \includegraphics[width=8.5cm]{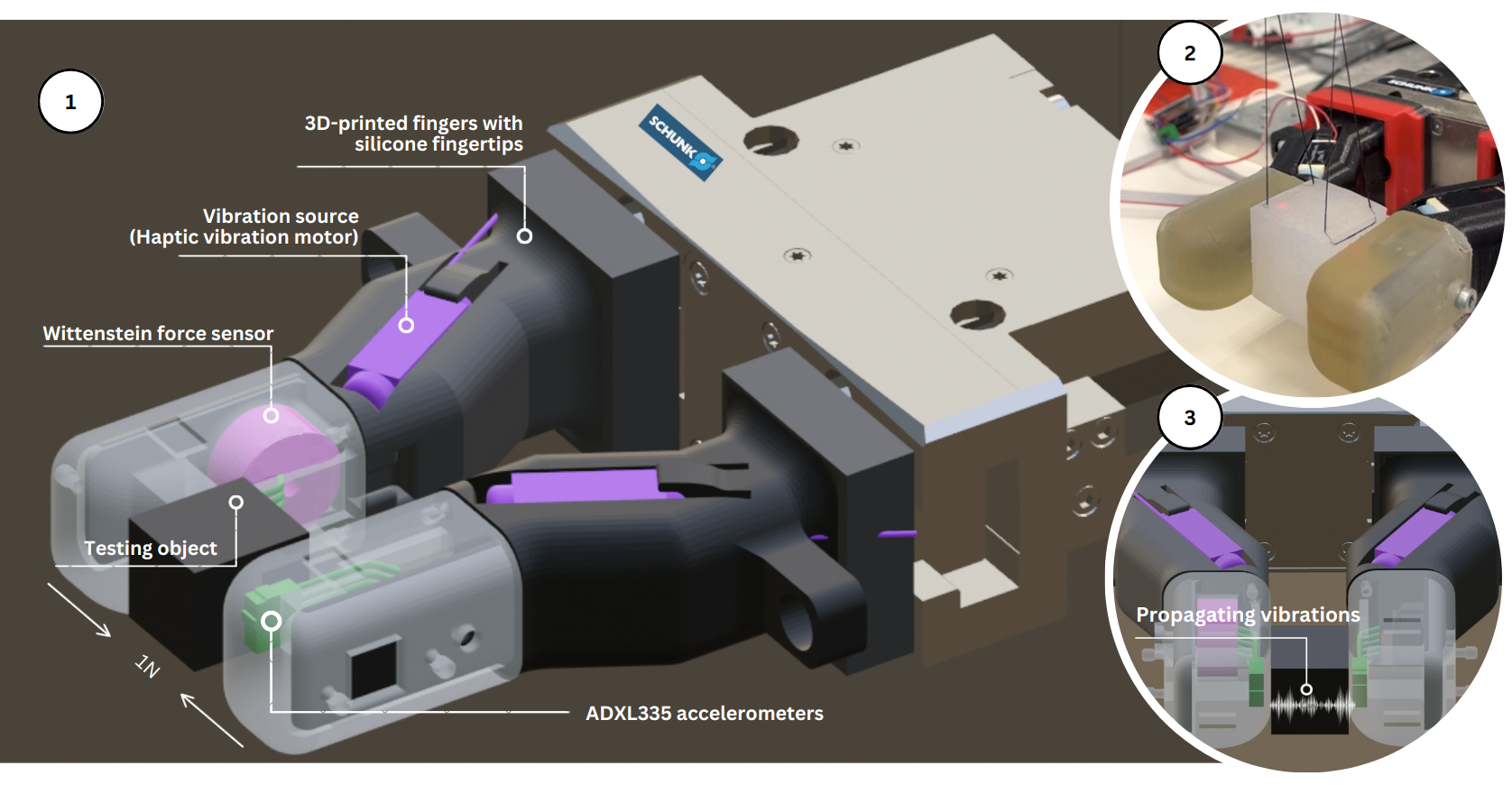}
    \caption{Representation of the Experimental Setup. (1) Schunk gripper with plastic fingers and silicone tips is squeezing a testing object. (2) Photo of the real setup. (3) Vibrations injected by a haptuator propagate through the object.}
    \label{fig:setup}
\end{figure}

\section{Related Works}
\subsection{Different Sensing Methods for Fine Manipulation}
Grasp control can be achieved via different sensing methods such as vision, audio, and touch sensing methods. Previous work offers a vision system that utilizes dual-camera setups to identify objects based on orientation, intensity, color, and texture \cite{doi:10.1177/0278364909346069}. Although some haptic sensors on fingertips enhance detail recognition, sensing typically occurs post-grasp, potentially leading to accidents. Vision systems are also expensive, computationally demanding, and suffer from occlusion issues, as stated in \cite{drimus2014design}, \cite{bhattacharjee2012haptic}, and \cite{mazhitov2023human}, making them unreliable for tasks requiring careful manipulation.

Acoustic perception is suitable for passive sensing, where objects generate sounds or resonate with the environment. Jiang and Smith's "Seashell Effect" exemplifies this, detecting objects through frequency resonance before grasping \cite{jiang2012seashell}. However, this method struggles to identify characteristics for optimal grasp configuration, necessitating direct contact for feature extraction. Similar limitations are seen in pre-touch sensing with optical proximity sensors \cite{hsiao2009reactive}, which are ineffective with transparent or reflective objects.

\subsection{Tactile Sensing}
Tactile sensing efficiently classifies soft and rigid objects, providing information about shape, texture, material, compliance, and temperature \cite{martinez2022review}, \cite{luo2017robotic}. A custom three-fingered robotic hand used soft tactile sensors for adaptive grip, translating pressure readings into voltage \cite{tsutsui2013robot}. However, pneumatic sensors might face challenges related to response time, accuracy, and scalability. 

Denei et al. developed a compact tactile sensor with capacitive pressure sensors, a microphone, and light sensors, enhancing texture classification and gripper control \cite{denei2015development}. Despite its compact size, the serial SPI interface likely caused signal delays and reduced sampling frequency. Increasing response speed might require a wider PCB for more ADC channels and parallel communication, compromising compactness.

Apart from the custom-made sensors, researchers often use high-end commercial tactile sensors such as BioTac, Weiss, PPS  RoboTouch, and Teksan \cite{kappassov2015tactile}, \cite{liu2017recent}. For example, Fishel and Loeb \cite{fishel2012bayesian} explored the use of Bayesian exploration techniques for intelligent texture identification. The authors propose a methodology that allows robots to effectively explore and identify different textures based on tactile feedback. Although they managed to classify 117 distinct textures, being a prohibitively expensive option for the majority of robotic systems, it does not address the practical constraints associated with deploying such technologies in real-world applications. 

\subsection{Object Stiffness Classification}
Stiffness classification is crucial for dexterous manipulation, providing information about an object's vulnerability to damage and slippage \cite{kappassov2015tactile}. Bhattacharjee, Rehg, and Kemp used a Cody robot's sensing forearm to differentiate objects based on stiffness and mobility, achieving 80\% accuracy \cite{bhattacharjee2012haptic}. This accuracy is increased in the case of utilizing a higher spatial resolution with more than 350 taxels. Based on these results, we can infer that tactile skin requires many sensors and, thus, other specialized hardware that can be difficult to prototype, which is also confirmed by \cite{ono2013touch}. 

Haptic gloves mimicking human skin offer another solution. For example, a fabric-based glove assisted Shadow Hand in discriminating objects of different stiffness \cite{gao2017wearable}. Piezoresistive materials, though flexible and sensitive, face issues with low spatial resolution, high hysteresis, and non-linearity \cite{qu2023recent}.

Park and Hwang's custom tactile modular sensor, embedded in a robotic fingertip, detected soft objects with high sensitivity, minimizing impact \cite{park2021softness}. This method avoids palpation, reducing the risk of damaging objects. Inspired by this, we used an active system to statically squeeze objects with minimal force for classification.

\subsection{Tactile Sensing using Vibro-Feedback}
Vibrations offer a low-cost, straightforward method to classify objects by tactile sensing \cite{gathmann2020wearable}. Active sensing involves applying force and generating vibrations to extract necessary data \cite{sandykbayeva2022vibrotouch}. Active vibration is a recent development in this field, enabling classification even when movement is restricted or objects are too small for sliding motions. Verification experiments showed classification rates comparable to those involving sliding motion.

\section{Methodology}
\subsection{Experimental Setup Description}
The main principle of active tactile sensing with vibro-feedback is the usage of a vibration source and two sensors recording vibration signals. We designed a system consisting of a Schunk ENG 100 robotic gripper with two 3D-printed plastic fingers covered with silicone fingertips. One finger has an embedded haptic vibration motor Haptuator Mark II and, thus, plays the role of an emitter of a vibration signal, whereas the second finger acts as a receiver. Both fingers have ADXL335 analog accelerometers to record vibrations. There is also a F/T sensor by Wittenstein HEX-21 in one of the fingertips utilized for force control.

We prepared eleven cube-shaped testing subjects with different stiffness parameters. They were made of SORTA-Clear™ silicone rubber 12, 18, and 40, FLEX plastic REC, Ultimaker TPU, and Ultimaker PLA of different infill densities: 0\%, 20\%, 40\%, 60\%, 80\%, 100\%. After squeezing each cube individually, the vibration signal propagated through the objects from the right (emitting) finger to the left (receiving) finger. Both accelerometers in their respective fingers recorded the signals for further analysis. 

\subsection{Experimental Procedure}
The gripper applied a static force of 1N to transmit the vibration signal through the testing cube. The STM32F4 microcontroller generated a chirp signal to the haptuator, recorded by the right accelerometer, and propagated through the cube to the left accelerometer. Sensor readings were saved and analyzed, revealing vibration absorption levels. Data from 50 trials ensured reproducibility, with denoising applied using a 50Hz uniform filter.

\begin{figure}[ht]
    \centering
    \includegraphics[width=6.5cm]{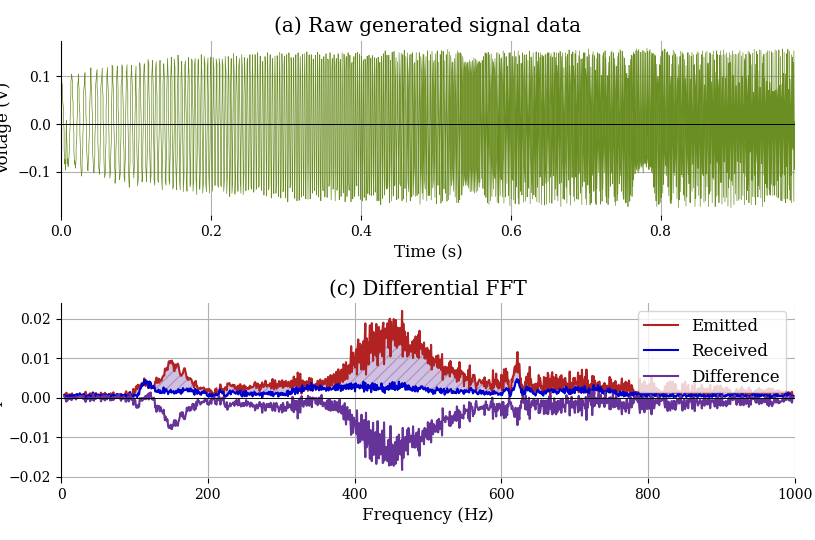}
    \caption{Signals used during the experiments: (a) Raw signal from the MCU; (b) Original FFT spectra of emitted and received signals and their difference.}
    \label{fig:signals}
\end{figure}

Fig. \ref{fig:signals} depicts the original signal generated by the MCU as compared to the signal received by the accelerometer on the emitting finger. The finger has absorbed frequencies from 200-400Hz due to the material properties and environmental effects as the system was not isolated completely from the table and the Schunk gripper. 
To gather more insights into the absorption properties, we computed differential FFT signals by subtracting the emitted signal values from the received signal ones (Fig. \ref{fig:signals}). The received differential FFT signals and classification based on these data confirmed the data shown in Table \ref{tab:youngs}.
\begin{figure}[ht]
    \centering
    \includegraphics[width=8.5cm]{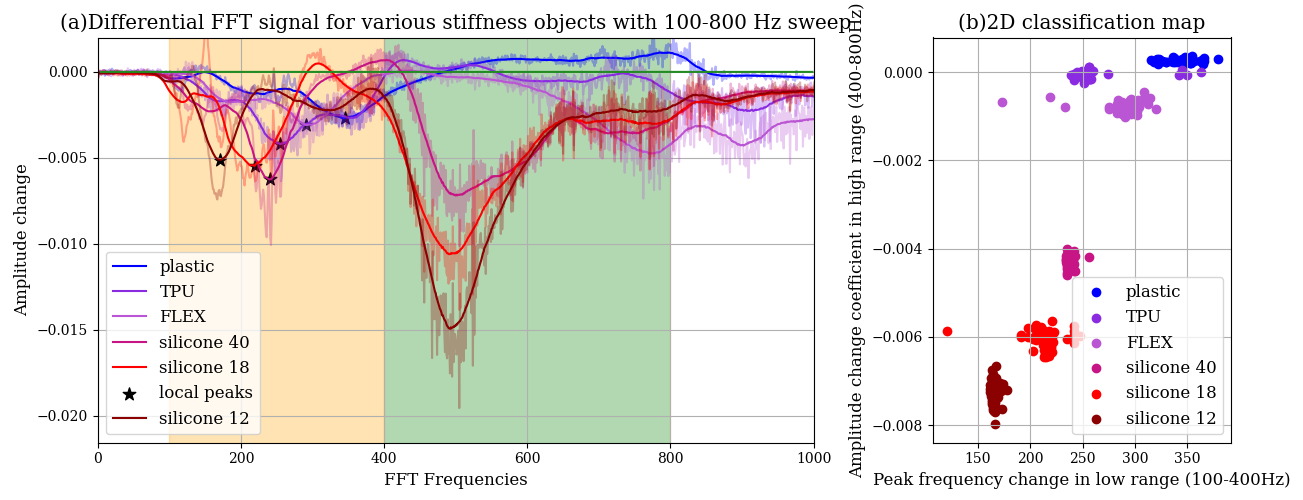}
    \caption{Classification results for variable stiffness: (a) FFT signal difference between emitter and receiver with low range peaks, and high range trends used for 2-feature classification; (b) Classification map}
    \label{fig:classification}
\end{figure}
\begin{figure}[ht]
    \centering
    \includegraphics[width=8.5cm]{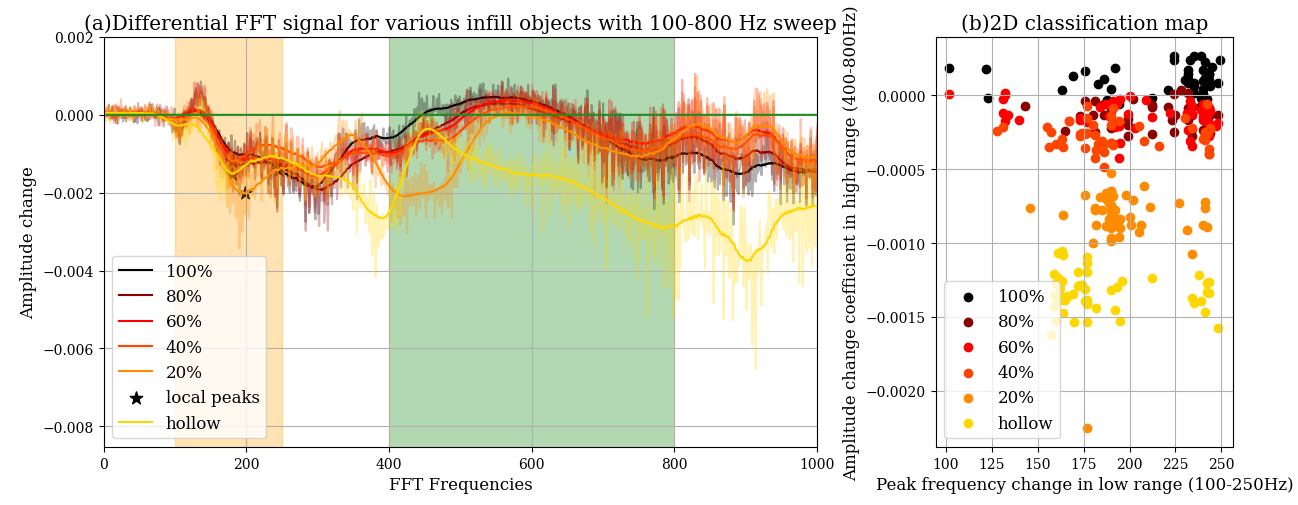}
    \caption{Classification results for variable infill: (a) FFT signal difference between emitter and receiver with low range peaks, and high range trends for 2-feature classification; (b) Classification map}
    \label{fig:classification2}
\end{figure}
\section{Discussion}
\subsection{Preliminary Experiments: Soft vs. Rigid}
In the preliminary experiments, the frequency sweep signals of 100-400 Hz, 100-600 Hz, and 100-800 Hz were utilized to analyze the differences in peak absorbance and amplification between soft and rigid objects. FFT Peak Absorbancy: At low frequencies (100-400 Hz), soft objects demonstrated a significant tendency to absorb the signal's FFT amplitude. Soft objects absorbed about 30-50\% of the peak amplitude, sometimes eliminating peak frequencies entirely. In contrast, plastic objects' absorbance is insignificant. FFT Peak Amplification: At higher frequencies (400-1000 Hz), plastic objects showed an increase in peak amplitude, amplifying the propagation and peak amplitudes, while soft objects exhibited an even stronger absorption than in the low-frequency region.
Using the identified differences in peaks and trends, a preliminary classification algorithm was developed. The final features used for classification are low-range peak changes and high-range trends. The high-range trends have been shown to exhibit increasing signal absorption with increasing object elasticity. For the low-range frequencies, peak changes on the contrary are happening at decreasing frequencies with increasing elasticity.\\
\begin{table}[ht]
    \centering
    \setlength{\tabcolsep}{4pt}
    \renewcommand{\arraystretch}{1.5} 
    \small 
    \begin{tabular}{cccccccccccc}
        Silicone12 & Silicone18 & Silicone40 & FLEX & TPU & PLA \\
        \hline
        0.4 & 0.664 & 1.696 & 63.7 & 67 & 3200 \\
    \end{tabular}
    \caption{Young's modulus for all materials used in the experiments in MPa.}
    \label{tab:youngs}
\end{table}
The classification maps summarize the collected results. As expected, based on the differential FFT signal results, the subjects made of different materials have fallen in their respective places on the 2D map for 2-feature classification (Fig. \ref{fig:classification}).
\subsection{Infill Classification} 
The second set of experiments involved using the same strategy for the identification of variable infills of the test objects (3D printed PLA).

The hollow subjects showed significant absorption of the emitted signal, while more solid cubes demonstrated less absorption and even amplification in the higher frequency range (450-600Hz) (Fig. \ref{fig:classification2}). The 2-feature classification map in Fig. \ref{fig:classification2} summarizes the plot of differential FFT signals. 
\section{Conclusion}
Our classification method, based on spectral analysis of frequency sweeps, successfully categorized objects by their material properties. This research highlights the potential of active vibro-feedback in tactile sensing for robotics. 

The research presents early preliminary work that can possess an increasing potential to discover more object properties that can be identified using a singular tactile sensing system. With that regard, there are a lot of unexplored areas of this approach, especially having active vibrational motors possibly affecting human interaction quality. Therefore future work would also include investigating opportunities to minimise the effect that the system might have on human interactions. Moreover, future research includes refining force control mechanisms, enhancing classification algorithms, and incorporating more object properties, paving the way for advanced tactile perception systems capable of nuanced object classification and manipulation.

\bibliographystyle{ieeetr}
\bibliography{bibliography} 




\end{document}